\documentclass{article}
\usepackage{spconf,amsmath,epsfig}
\usepackage{xcolor,subfigure}
\usepackage{amsmath}
\usepackage{amssymb}
\usepackage{amsthm,bm}
\usepackage{algorithm, algorithmic}

\def\mr{\mathrm}
\def\mb{\mathbf}
\def\mbb{\mathbb}

\begin{document}\sloppy

\def\x{{\mathbf x}}
\def\L{{\cal L}}

\title{SR-GAN: Semantic Rectifying Generative Adversarial Network for Zero-shot Learning}
%
\name{Zihan Ye$^{1,5}$$^\star$, Fan Lyu$^{1, 2}$$^\star$\thanks{$\star$The first two authors contributed to this work equally.}, Linyan Li$^{3}$, Qiming Fu$^{1,6}$, Jinchang Ren$^{4}$, Fuyuan Hu$^{1,7}$$^{\star \star}$\thanks{$\star \star$Corresponding author: fuyuanhu@mail.usts.edu.cn.}}
\address{$^{1}$Suzhou University of Science and Technology, China\\$^{2}$Tianjin University, China\\$^{3}$Suzhou Institute of Trade \& Commerce, China\\$^{4}$University of Strathclyde, UK\\$^{5}$Virtual Reality Key Laboratory of Intelligent Interaction and Application Technology of Suzhou, China\\$^{6}$Key Laboratory of Intelligent Building Energy Efficiency, China\\$^{7}$Suzhou Key Laboratory for Big Data and Information Service, China}

\maketitle

\begin{abstract}
	The existing Zero-Shot learning (ZSL) methods may suffer from the vague class attributes that are highly overlapped for different classes. Unlike these methods that ignore the discrimination among classes, in this paper, we propose to classify unseen image by rectifying the semantic space guided by the visual space. First, we pre-train a Semantic Rectifying Network (SRN) to rectify semantic space with a semantic loss and a rectifying loss. Then, a Semantic Rectifying Generative Adversarial Network (SR-GAN) is built to generate plausible visual feature of unseen class from both semantic feature and rectified semantic feature. To guarantee the effectiveness of rectified semantic features and synthetic visual features, a pre-reconstruction and a post reconstruction networks are proposed, which keep the consistency between visual feature and semantic feature. Experimental results demonstrate that our approach significantly outperforms the state-of-the-arts on four benchmark datasets.
\end{abstract}
\section{Introduction}
\label{sec:intro}

The classical pattern of object recognition classifies image into categories only seen in training stage~\cite{lyu2018coarse,Lyu2019attend}.
In contrast, zero-shot learning (ZSL) aims at exploring unseen image categories, which gets a lot of attention~\cite{lampert2014attribute, xian2017zero, akata2016label,lampert2009learning,reed2016learning, zhu2018generative,li2018discriminative, song2018selective, jiang2018learning} in recent years.
By using the intermediate semantic features (obtained from human-defined attribute) of both seen and unseen classes, the previous methods inference unseen classes of image.
However, the human-defined attributes are highly overlapped for similar classes, which is prone to failure prediction.
In this paper, we propose a Semantic Rectifying Network (SRN) to make semantic feature more distinguishable, and a Semantic Rectifying Generative Adversarial Network to synthesize unseen classes data from both corresponding semantic and rectified semantic feature. By synthesizing missing features, the unseen classes can be classified by supervised classification approach, like nearest neighbors.

ZSL is challenging~\cite{xian2017zero}, because the images to be predicted are from unseen classes. 
\cite{chao2016empirical} proposes the generalized zero-shot learning (GZSL) to improve the expandability of ZSL. 
Different from ZSL, GZSL also has seen classes used at test time.
Some studies~\cite{lampert2014attribute,frome2013devise,romera2015embarrassingly,akata2016label,xian2016latent,kodirov2017semantic} project image feature to semantic space and treat image as the class with closest semantic feature to the projected image feature.  
Recently, inspired by the generative ability of generative adversarial networks (GANs), ~\cite{goodfellow2014generative,zhu2018generative, zihan2018DAU} leverages GANs to generate synthesized visual feature from semantic features and noise samples, and designs a visual pivot regularization to simulate visual distribution with greater inter-class discrimination.
By generating missing features for unseen classes, they convert ZSL to a conventional classification problem, and some classical method such as nearest neighbors can be used.

\begin{figure}[t]
	\centering
	\includegraphics[width=1.05\columnwidth]{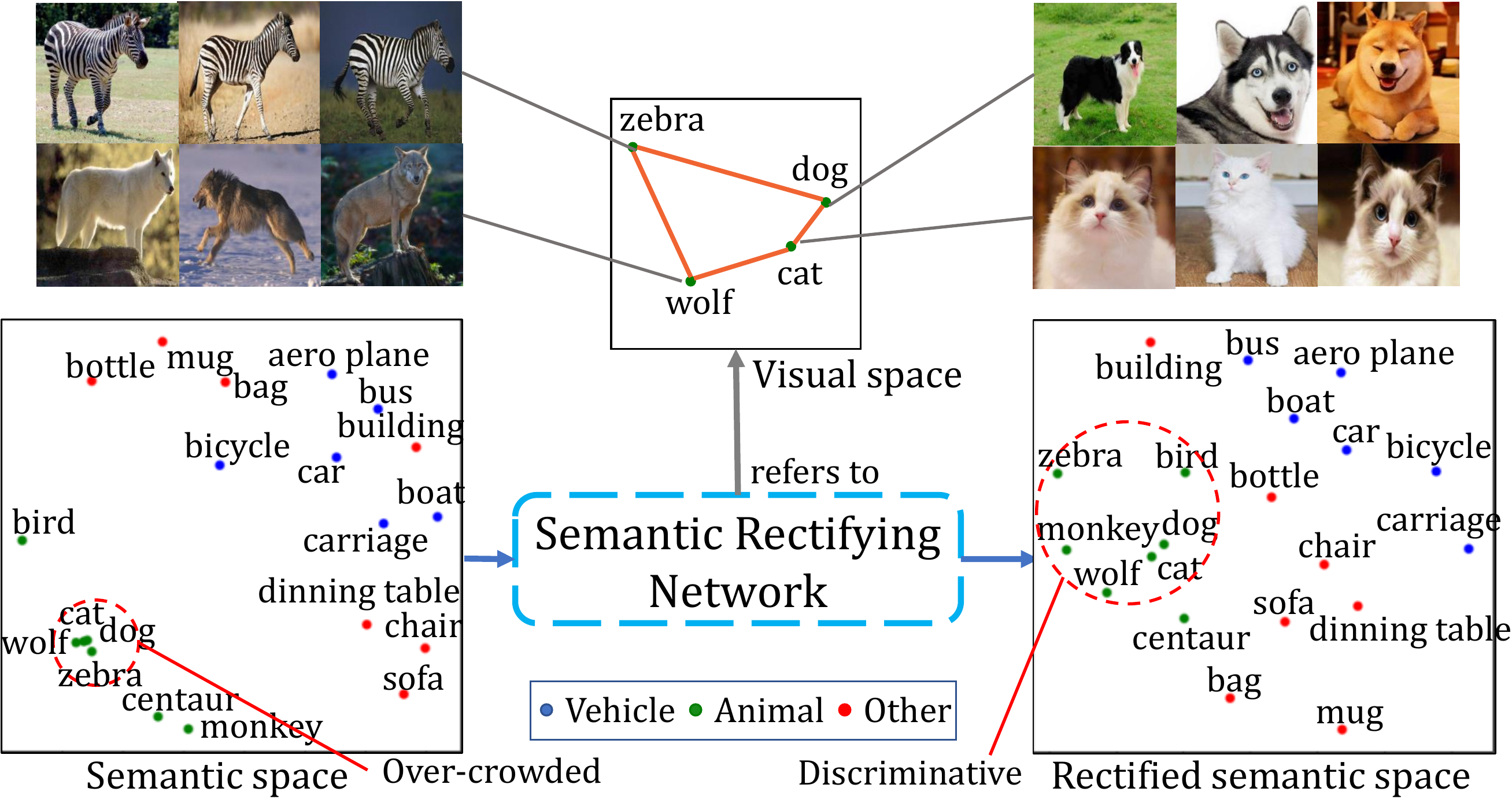}
	\caption{Schema of the proposed method. We visualize semantic features and corresponding rectified semantic features of 20 classes from APY by multidimensional scaling (MDS) ~\cite{kruskal1964multidimensional}. Classes overlapped in semantic space can be rectified by the proposed semantic rectifying network. }
	\label{fig:one}
	\vspace{-15px}
\end{figure}

\begin{figure*}[t]
	\centering
	\includegraphics[width=0.8\linewidth]{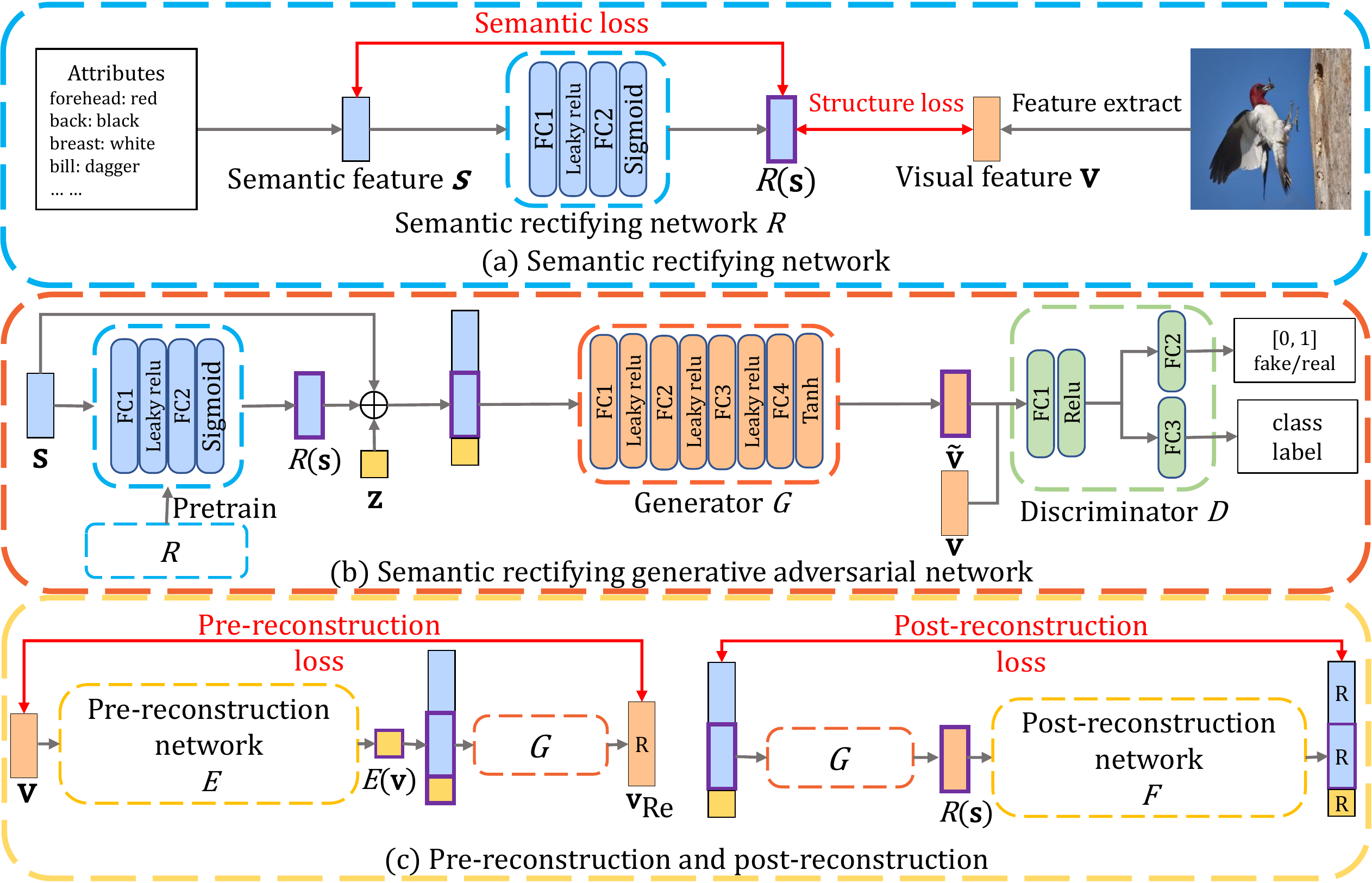}
	\caption{The architecture of Semantic Rectifying Generative Adversarial Network. 
		(a) Semantic Rectifying Network (SRN) rectifies semantic feature into a more discriminative space guided by the seen visual information. 
		(b) Semantic Rectifying Generative Adversarial Network makes use of the rectified semantic feature to generate visual feature, and discriminate which into real of fake.
		(c) the Pre-reconstruction Network allows our generator to learn a more exact distribution of visual features; the Post-reconstruction reconstruction network allows synthetic visual feature to be translated into their semantic features.}
	\label{fig:two}
	\vspace{-15px}
\end{figure*}

However, semantic features are difficult to exactly define due to the overlap of the common properties such as color, shape, and texture for many class. 
For example, elephant and tiger are both giant and with tail.
Thus, as shown in Fig.~\ref{fig:one}, semantic features of some similar classes may cluster in a small region and are indistinguishable in the semantic space. 
Obviously, wolf, cat, dog and even zebra are quite close in semantic space because these classes have overlapped attributes or similar descriptions. 
Accordingly, synthesizing visual feature from these indistinguishable semantic features is unreasonable.
Some works seek to address this problem by learning an extra space apart from visual space.
\cite{li2018discriminative} and \cite{song2018selective} propose to automatically mine latent and discriminative semantic feature from visual feature.
\cite{jiang2018learning} constructs an align space as a trade-off between semantic and visual space.
However, as containing two heterogeneous kinds of information, the aligned space may be misled by the noise from image.

In this paper, we adopt another strategy, i.e., to rectify the semantic space into a more reasonable one guided by visual feature.
We first design a Semantic Rectifying Network (SRN) to pre-rectify undistinguished semantic feature.
Based on the generative adversarial network, as shown in Fig.~\ref{fig:two}, a Semantic Rectifying Generative Adversarial Network (SR-GAN) is then proposed to generate visual feature from the rectified semantic feature.
As shown in Fig.~\ref{fig:one}, semantic features that are over-crowded in original feature space become distinguishable in the rectified semantic space. 
Moreover, a pre-reconstruction and a post-reconstruction network are proposed, which construct two cycles to preserve the consistency of semantic feature and visual feature.
We evaluate the proposed method on four datasets, i.e., AWA1, CUB, APY and SUN.
The experimental results demonstrate that our approach outperforms state-of-the-art for both ZSL and GZSL tasks.

\section{Methodology}
\subsection{Formulation}
\vspace{-5px}
Given an image $I$, the proposed model can recognize it as a specific class $c$ even that is unseen.
Following~\cite{xian2017zero}, we take instance $\{\mb{v}, \mb{s}^\mr{s}, c^\mr{s}\}$ as input in the training stage, where $\mb{v}$ is the visual feature of $I$ in the visual space $\mathcal{V}$, $\mb{s}^\mr{s}$ in the semantic space $\mathcal{S}$ is the semantic features extracted from attributes or descriptions of class, and $c^\mr{s}$ denotes the corresponding seen class label. $\mbb{C}^{s}$ is the set of seen class labels.
In the testing stage, given an image, ZSL and GZSL will recognize it as an unseen class $c^\mr{u}$ or an either seen or unseen class $c^\mr{s+u}$.
As shown in Fig~\ref{fig:two}, our framework consists of three components: (a) Structure Rectifying Network (SRN) ${R}$ to rectify semantic space with refers visual space; (b) Semantic Rectifying Generative Adversarial Network (SR-GAN) to synthesize pseudo visual feature and do zero-shot classification; (c) a pre-reconstruction and a post-reconstruction networks to keep the consistency between visual feature and semantic feature. 

\subsection{Semantic rectifying network}
\vspace{-5px}
The primary obstacle of ZSL is that difficult to guarantee the distribution of visual space and semantic space are corresponding.
Specifically, the vague class attributes and descriptions make model confusing, as well as generate convincing visual feature.
To this end, we design a Semantic Rectifying Network (SRN), denoted as $R$ and shown in Fig.~\ref{fig:two}(a), to rectify the class structures between the visual space and the semantic space. 
SRN consists of a multi-layer perceptron (MLP) activated by Leaky ReLU~\cite{maas2013rectifier}, and the output layer has a Sigmoid activation. We define the visual pivot vector $\mb{p}$ for each class across dataset, and for the $c$-th class we have
\begin{eqnarray}
\mb{p}_{c} = \frac{1}{N_{c}}\sum_{i=1}^{N_{c}}\mb{v}_{i}^{c},
\end{eqnarray}
where $N_{c}$ is the number of instances with class $c$, and $\mb{v}_{i}^{c}$ is the $i$-th visual feature of class $c$. 
We argue that for any two classes, their visual pivots have similar relationship as their semantic feature.
Thus, we use the cosine similarity function $\delta$ to provide the similarities between pair of visual pivot and semantic feature, and obtain a rectifying loss for SRN
\begin{align}
\label{eq:lr}
L_\text{R} =& \frac{1}{|\mbb{C}^{S}|^{2}} \sum_{i=1}^{|\mbb{C}^{S}|}\sum_{j=1}^{|\mbb{C}^{S}|}\|\delta(\mb{p}_{i}, \mb{p}_{j}) - \delta(R(\mb{s}_{i}),R(\mb{s}_{j}))\|_{2} \nonumber \\
&+ \mathbf{E}_{\mb{s} \sim p_{\mb{s}}}\|\mb{s} - R(\mb{s})\|_{2}, 
\end{align}
where $|\mbb{C}^{s}|$ is the number of seen classes.
The first term of Eq.~(\ref{eq:lr}) is the structure loss expressing the directional distance between the rectified semantic features and visual features, and the second term is a semantic loss, which measures the consumption of semantic information after rectifying. 
Note that we fix the parameters of SRN after training it.
\begin{algorithm}[t]
	\renewcommand{\algorithmicrequire}{\textbf{Input:}}
	\renewcommand{\algorithmicensure}{\textbf{Output:}}
	\caption{Training procedure of our approach.}
	\label{alg:srgan}
	\begin{algorithmic}[1]
		\REQUIRE The batch size $m=1024$, learning rate $\lambda_{l} = 0.0001$, the number of discriminator training loop $n_{d}=5$.
		\STATE Initialize $R$ randomly
		\FOR{SRN training iterations}
		\STATE Sample seen visual features $\mb{v}$ and semantic features $\mb{s}^\mr{s}$
		\STATE Update $R$ by Eq.~\ref{eq:lr}
		\ENDFOR
		\STATE Fix $R$ and initialize $G, D, F, E$ randomly
		\FOR{SR-GAN training iterations}
		\FOR{$n_{d}$}
		\STATE Sample a mini-batch of visual features $\mb{v}$, corresponding semantic features $\mb{s}$ and random vector $\mb{z}$
		\STATE Fix $G,E,F$ and update $D$ by Eq.~\ref{eq:ld}
		\ENDFOR
		\STATE Sample a mini-batch of visual features $v$, corresponding semantic features $s$ and random vector $z$
		\STATE Fix $D$ and update $G$ by Eq.~\ref{eq:lg}
		%
		\STATE Update $G, E$ by Eq.~\ref{eq:le}
		%
		\STATE Update $G, F$ by Eq.~\ref{eq:lre}
		\ENDFOR
	\end{algorithmic}  
\end{algorithm}

\subsection{Semantic rectifying GAN}
\vspace{-5px}

Generative Adversarial Network (GAN) has been demonstrated useful for ZSL \cite{zhu2018generative}, as the ability to generate visual features from semantic feature.
However, indiscriminately feeding vague semantic feature into a generator may undermine the generated visual feature.
By a pre-trained SRN model, we can easily obtain more distinguished semantic feature.
Therefore, we design a semantic rectifying GAN (SR-GAN) model that translates these rectified semantic features into visual features.
As shown in Fig.~\ref{fig:two}(b), the proposed SR-GAN has a generator $G$ and a discriminator $D$.
For $G$, we have three types of input, i.e., the original semantic feature $\mb{s}$, the rectified semantic feature $R(\mb{s})$, and the random vector $\mb{z}$ sampled from the normal distribution.
$G$ consists of a four-layers MLP with a residual connection, where the first three layers are with leaky ReLU activation and the output layer is activated by Tanh.
The loss of generator is defined as: 
\begin{equation}
\label{eq:lg}
L_\text{G} =-\mathbf{E}_{\mb{z}\sim p_{\mb{z}}}\left[D(G(\mb{s}, R(\mb{s}),\mb{z}))\right] + L_{\text{cls}}(G(\mb{s}, R(\mb{s}), \mb{z})) + L_{\text{VP}},
\end{equation}
where $\mathbf{E}\left[\cdot\right]$ denotes the expected value.
The first term of Eq.~(\ref{eq:lg}) is a standard generator loss of Wasserstein GAN (W-GAN)~\cite{arjovsky2017wasserstein}. 
The second term is a cross entropy loss and the third term is a visual pivot loss. 
A visual pivot loss can be computed as the Euclidean distance between the prototypes of synthesized features and real features for each class:
\begin{eqnarray}
\label{eq:lvp}
L_\text{VP} = \frac{1}{|\mbb{C}^{s}|}\sum_{i=1}^{|\mbb{C}^{s}|}\left\| \mb{p}_{i} - \frac{1}{N_{i}}\sum_{j=1}^{N_{i}} G(\mb{s}_{i},R({\mb{s}}_{i}),\mb{z}_{j}) \right\|_2.
\end{eqnarray}
For the discriminator $D$, it takes synthetic features or real features as input and has two output branches. 
One branch is to distinguish the input is real or fake, and the other branch will classify the given input into different classes wherever in ZSL or GZSL.
Consequently, the loss of $D$ is defined as:
\begin{align}
\label{eq:ld}
L_\text{D} =& \mathbf{E}_{\mb{z}\sim p_{\mb{z}}}\left[ D(G(\mb{s},R(\mb{s}),\mb{z})) \right] - \mathbf{E}_{\mb{v} \sim p_{data}}\left[ D(\mb{v}) \right] \nonumber \\
&+ \lambda(\| \bigtriangledown_{\widehat{\mb{v}}}D(\widehat{\mb{v}}) \|_{2} -1)^{2}\\
&+ \frac{1}{2}(\text{L}_{\text{cls}}(G(\mb{s},R(\mb{s}),\mb{z})) + \text{L}_{\text{cls}}(\mb{v}))  \nonumber,
\end{align}
where the first second terms are the standard discriminator loss of W-GAN, and the third term is the gradient penalty. 
This gradient penalty term do help Wasserstein GAN get rid of pathological behavior~\cite{arjovsky2017wasserstein}, and $\lambda$ denotes the penalty coefficient.
The last term is an auxiliary classification loss.

\begin{table*}[t]
	\begin{center}
		\centering
		\caption{Comparison with the state-of-the-art method on four datasets.} \label{tab:three}
		\resizebox{.9\linewidth}{!}{
			\begin{tabular}{p{1.8cm}|p{0.6cm}p{0.5cm}p{0.5cm}p{0.6cm}|p{0.5cm}p{0.5cm}p{0.6cm}|p{0.5cm}p{0.5cm}p{0.6cm}|p{0.5cm}p{0.5cm}p{0.6cm}|p{0.5cm}p{0.5cm}p{0.6cm}}
				\hline
				\multicolumn{1}{c}{$ $}&\multicolumn{4}{c|}{Zero-Shot Learning}& \multicolumn{12}{c}{Generalized Zero-Shot Learning}\\
				\hline
				$ $&AWA1&CUB&APY&SUN  & \multicolumn{3}{|c|}{AWA1} & \multicolumn{3}{|c|}{CUB} & \multicolumn{3}{|c|}{APY}& \multicolumn{3}{|c}{SUN}\\
				\hline
				Classifier&$T1$&$T1$&$T1$&$T1$&$U$&$S$&$H$&$U$&$S$&$H$&$U$&$S$&$H$&$U$ &$S$&$H$\\
				\hline
				DAP~\cite{lampert2014attribute}&44.1&40.0&33.8&39.9&0&\textbf{88.7}&0&1.7&67.9&3.3&4.8&78.3&9.0&4.2&25.1&7.2\\
				
				CONSE~\cite{norouzi2013zero}&45.6&34.3&26.9&38.8&0.4&88.6&0.8&1.6&\textbf{72.2}&3.1&0&\textbf{91.2}&0&6.8&39.9&11.6\\
				
				SSE~\cite{zhang2015zero}&60.1&43.9&34.0&51.5&7.0&80.5&12.9&8.5&46.9&14.4&0.2&78.9&0.4&2.1&36.4&4.0\\
				
				DEVISE~\cite{frome2013devise}&54.2&52.0&39.8&56.5&13.4&68.7&22.4&23.8&53.0&32.8&4.9&76.9&9.2&16.9&27.4&20.9\\
				
				SJE~\cite{akata2015evaluation}&65.6&53.9&32.9&53.7&11.3&74.6&19.6&23.5&59.2&33.6&3.7&55.7&6.9&14.7&30.5&19.8\\
				
				LATEM~\cite{xian2016latent}&55.1&49.3&35.2&55.3&7.3&71.7&13.3&15.2&57.3&24.0&0.1&73.0&0.2&14.7&28.8&19.5\\
				
				ESZSL~\cite{romera2015embarrassingly}&58.2&53.9&38.3&54.5&6.6&75.6&12.1&12.6&63.8&21.0&2.4&70.1&4.6&11.0&27.9&15.8\\
				
				ALE~\cite{akata2016label}&59.9&54.9&39.7&58.1&16.8&76.1&27.5&23.7&62.8&34.4&4.6&73.7&8.7&21.8&33.1&26.3\\
				
				SYNC~\cite{changpinyo2016synthesized}&54.0&55.6&23.9&56.3&8.9&87.3&16.2&11.5&70.9&19.8&7.4&66.3&13.3&7.9&\textbf{43.3}&13.4\\
				
				SAE~\cite{kodirov2017semantic}&53.0&33.3&8.3&40.3&1.8&77.1&3.5&7.8&54.0&13.6&0.4&80.9&0.9&8.8&18.0&11.8\\
				
				GAZSL~\cite{zhu2018generative}&68.2&55.8&41.13&61.3&19.2&86.5&31.4&23.9&60.6&34.3&14.17&78.63&24.01&21.7&34.5&26.7\\
				PSR~\cite{annadani2018preserving}&-&\textbf{56.0}&38.4&61.4&-&-&-&24.6&54.3&33.9&13.5&51.4&21.4&20.8&37.2&26.7\\
				
				CDL~\cite{jiang2018learning}&69.9&54.5&43.0&\textbf{63.6}&28.1&73.5&40.6&23.5&55.2&32.9&19.8&48.6&28.1&21.5&34.7&26.5\\
				
				\hline
				SR-GAN&\textbf{71.97}&55.44&\textbf{44.02}&62.29&\textbf{41.46}&83.08&\textbf{55.31}&\textbf{31.29}&60.87&\textbf{41.34}&\textbf{22.34}&78.35&\textbf{34.77}&\textbf{22.08}&38.29&\textbf{27.36}\\
				\hline
			\end{tabular}
		}
	\end{center}
	\vspace{-25px}
\end{table*}

\begin{table*}[t]
\begin{center}
	\centering
	\caption{Effects of different components on four datasets.} \label{tab:ablation}
	\resizebox{.95\linewidth}{!}{
		\begin{tabular}{p{4.2cm}|p{0.6cm}p{0.5cm}p{0.5cm}p{0.6cm}|p{0.5cm}p{0.5cm}p{0.6cm}|p{0.5cm}p{0.5cm}p{0.6cm}|p{0.5cm}p{0.5cm}p{0.6cm}|p{0.5cm}p{0.5cm}p{0.6cm}}
			\hline
			\multicolumn{1}{c}{$ $}&\multicolumn{4}{c|}{Zero-Shot Learning}& \multicolumn{12}{c}{Generalized Zero-Shot Learning}\\
			\hline
			$ $&AWA1&CUB&APY&SUN & \multicolumn{3}{|c|}{AWA1} & \multicolumn{3}{|c|}{CUB} & \multicolumn{3}{|c|}{APY}& \multicolumn{3}{|c}{SUN}\\
			\hline
			Classifier&$T1$&$T1$&$T1$&$T1$&$U$&$S$&$H$&$U$&$S$&$H$&$U$&$S$&$H$&$U$ &$S$&$H$\\
			\hline
			SR-GAN:baseline&69.13&52.88&39.98&60.49&34.04&\textbf{84.29}&48.48&27.53&58.96&37.53&19.39&77.70&31.03&21.46&36.16&26.93\\
			
			SR-GAN:baseline+rec&69.11&53.14&40.79&61.18&35.05&83.41&49.36&27.96&\textbf{61.92}&38.53&20.38&\textbf{79.80}&32.47&21.88&38.26&27.83\\
			
			SR-GAN:baseline+SRN&70.43&55.21&43.44&61.25&37.92&83.84&52.22&30.45&61.10&40.64&21.65&72.39&33.33&20.69&\textbf{39.46}&27.15\\
			
			SR-GAN:baseline+rec+SRN&\textbf{71.97}&\textbf{55.44}&\textbf{44.02}&\textbf{62.29}&\textbf{41.46}&83.08&\textbf{55.31}&\textbf{31.29}&60.87&\textbf{41.34}&\textbf{22.34}&78.35&\textbf{34.77}&\textbf{22.08}&38.29&\textbf{27.36}\\
			\hline
		\end{tabular}
	}
\end{center}
\vspace{-25px}
\end{table*}

\subsection{Pre-reconstruction and post-reconstruction}
\vspace{-5px}

By the above process, the model is able to synthesize good visual feature to some extent.
However, there still exists a significant problems, i.e., the generated visual feature has poor consistency with the input semantic.
Accordingly, we propose a pre-reconstruction and a post-reconstruction modules to keep the consistency of visual feature and semantic feature.
Specifically, as shown in Fig.~\ref{fig:two}(c), the pre-reconstruction network $E$ takes the real visual feature $\mb{v}$ as input, but the constructed semantic feature and random noise are then fed into the generator $G$, and builds consistency loss between visual feature and reconstructed visual feature.
On the contrary, the post-reconstruction network $F$, keeping in step with the generator $G$, takes the generated feature $\tilde{\mb{v}}$ as input, and builds consistency loss between reconstructed semantic feature and random noise.
The pre-reconstruction loss for $E$ and the post-reconstruction loss for $F$ can be computed by
\begin{equation}
\label{eq:le}
L_\text{E} = \|G(\mb{s}, R(\mb{s}), E(\mb{v})) - \mb{v}\|_{1},
\end{equation}
\begin{equation}
\label{eq:lre}
L_\text{F} = \|F(G(\mb{s}, R(\mb{s}), \mb{z})) - [\mb{s}, R(\mb{s}),  \mb{z}]\|_{1}.
\end{equation}
where $[\cdot]$ represents the concatenation operator.
Obviously, $E(G(\cdot))$ and $G(F(\cdot))$ can be considered as two Auto-Encoders~\cite{kingma2014auto}. 
The pre-reconstruction $E\circ G$ allows generator learn a more convincing visual distribution by forcing generator to restore real visual feature $\mb{v}$ from encoded random vector $E(G(\cdot))$.
And the post-reconstruction $G(F(\cdot))$ enhances the relationship between synthetic visual feature and the corresponding class semantics by minimizing the difference between the reconstructed and original semantic feature.
Finally, by integrating the pre-reconstruction loss and post-reconstruction loss, the loss of $G$ can be modified as :
\begin{align}
\text{L}_\text{G} =&-\mathbf{E}_{\mb{z}\sim p_{\mb{z}}}\left[D(G(\mb{s}, R(\mb{s}), \mb{z}))\right] \nonumber \\
&+ L_{\text{cls}}(G(\mb{s},R(\mb{s}),\mb{z})) + L_{\text{VP}} + L_{\text{E}} + L_{\text{F}}.
\end{align}
We have the training procedure in Algorithm \ref{alg:srgan}.
We first train SRN, and fix its parameter after training.
Then we train generator and discriminator of SR-GAN in turn, but train discriminator more times (default value is 5, following ~\cite{arjovsky2017wasserstein}) than generator.
Note that we experientially update the parameters of $G$ several times for Eq.~\ref{eq:lg}, Eq.~\ref{eq:le} and Eq.~\ref{eq:lre}, as we find this is very useful to obtain a more reliable generator.

\section{Experiment}

\subsection{Implemented details}
\vspace{-5px}
\noindent\textbf{Datasets}.
We evaluate our approach on four benchmark datasets for ZSL and GZSL: 
(1) \textit{Caltech-UCSD-Birds 200-2011} (CUB)~\cite{wah2011caltech} has 11,788 images, 200 classes of birds annotated with 312 attributes; 
(2) \textit{Animals with Attributes} (AWA)~\cite{lampert2009learning} is coarse-grained and has 30,475 images, 50 types, and 85 attributes; 
(3) \textit{Attribute Pascal and Yahoo} (APY)~\cite{farhadi2009describing} contains 15,339 images, 32 classes and 64 attributes; 
(4) \textit{SUN Attribute} (SUN)~\cite{patterson2014sun} annotates 102 attributes on 14,340 images from 717 types of scene. 
For all four datasets, we use the widely-used ZSL and GZSL split proposed in ~\cite{xian2017zero}.

\noindent\textbf{Evaluation metrics}.
We use the evaluation metrics proposed in ~\cite{xian2017zero}. 
For ZSL, we measure the average per-class top-1 accuracy (T1) of unseen classes $\mbb{C}^{u}$. 
For GZSL, we compute the average per-class top-1 accuracy of seen classes $\mbb{C}^{s}$, denoted by $S$, and unseen classes $\mbb{C}^{u}$, denoted by $U$, and their harmonic mean, i.e. $H = 2 \times (S \times U)/(S + U)$.

\begin{figure}[t]
	\centering
	\includegraphics[width=.95\linewidth]{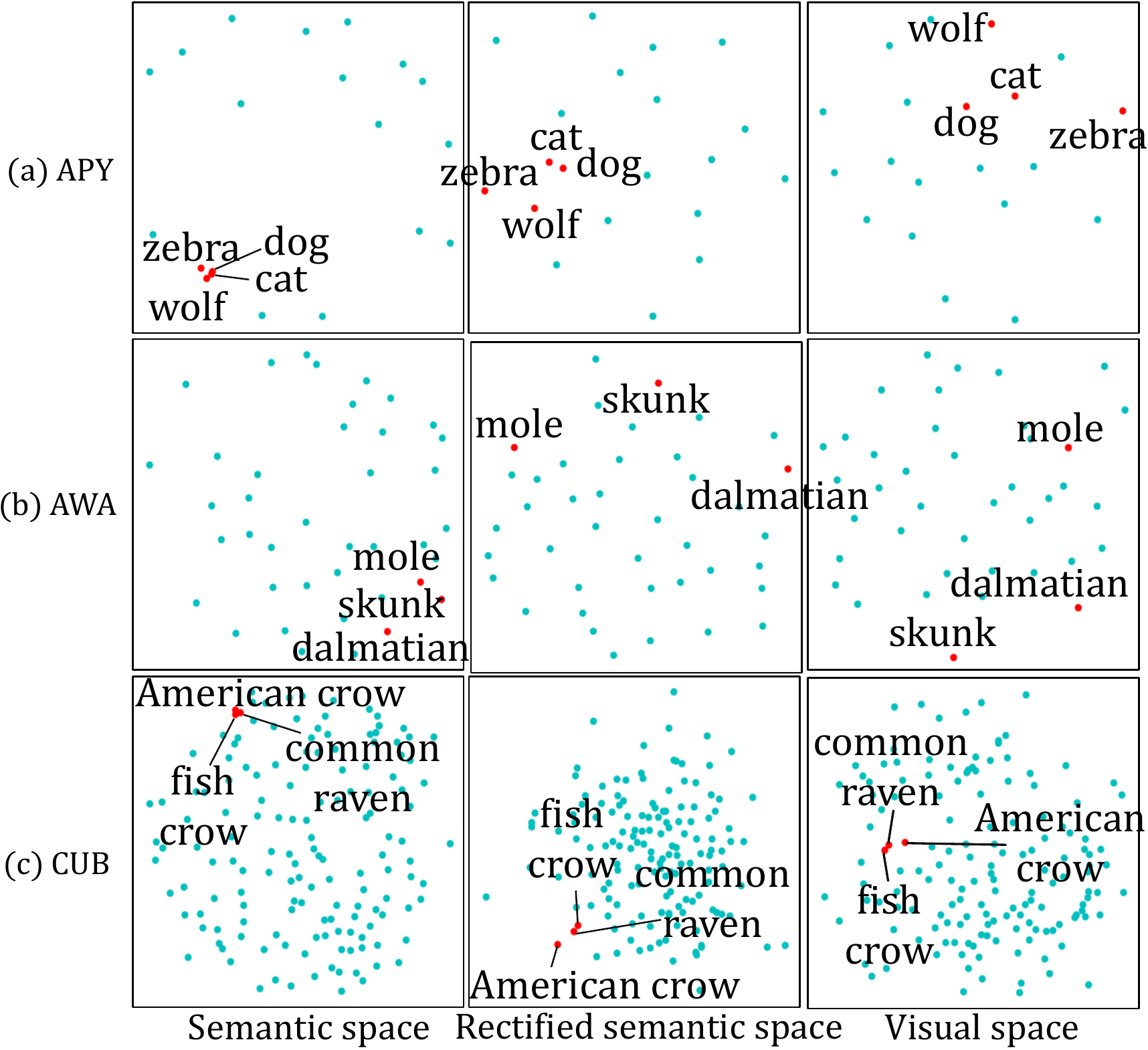}
	\caption{Seen class visualizations in APY, AWA1, and CUB.}
	\label{fig:mds}
	\vspace{-15px}
\end{figure}

\subsection{Comparison to the state-of-the-arts}
\vspace{-5px}
We compare the proposed method against several state-of-the-art methods in both ZSL and GZSL setting in Table~\ref{tab:three}. 
Obviously, SR-GAN achieves the best performance in most situations.
On the one hand, for ZSL learning, we get two new state-of-the-arts on AWA1 and APY, and also obtain comparable results on the other datasets.
On the other hand, for GZSL learning, our approach achieve the best performance for $U$ and $H$ on all four datasets, which indicates that our approach can improve the performances of unseen classes while maintaining a high accuracy of seen classes $S$. 
Specifically, for $U$ we have a great boost for all datasets, which indicates our approach alleviates the seen-unseen bias problem better than other approaches under the GZSL scenario.
Unlike other methods that utilize fixed semantic information, SR-GAN make original semantic feature more discriminative, which is similar to CDL~\cite{jiang2018learning}. 
However, the aligned space in CDL is a compromise between semantic and visual space. 
In addition, CDL computes the feature similarities in semantic space, visual space and aligned space separately, which reduce the final performance. 
And our approach automatically explores the discrimination in rectifying space and also preserve the original semantic information to some extent.

\subsection{Ablation study}
\vspace{-5px}
\label{sec:ablation}
We analyze the importance of every component of the proposed framework in Table~\ref{tab:ablation}. We denote two reconstruction networks, semantic rectifying network and the rest of our whole model as rec, SRN, and baseline, respectively. We evaluate three variants of our model by removing different components.
The performance of the baseline is unremarkable for almost all datasets except $S$ of AWA1. 
With the help of semantic rectifying network (rec), the performances slightly increases, e.g. for $U$ in AWA1, SR-GAN:baseline+rec is better than baseline only (35.05\%vs. 34.04\%). 
It indicates that our reconstruction networks enhance the imagination of our model for unseen classes. 
With the SRN, the performances significantly increases: the performance of SR-GAN:baseline+SRN in all dataset for almost all accuracies is better than the baseline, which indicates that our SRN effectively rectifies semantic features to a more distinguishable space to generate more realized visual features.

%

\subsection{Visualization of rectifying space}
\vspace{-5px}
To validate that the proposed SRN is effective for rectifying semantic features more distinguishable, we visualize all seen classes of APY, AWA1 and CUB and visualize their semantic features, the corresponding rectified semantic features, and the pivots of visual features in a 2-D plane by multidimensional scaling (MDS) ~\cite{kruskal1964multidimensional}. 
Visualization results are shown in Figure~\ref{fig:mds}. 
We can see that the original semantic features are not distinguishable enough, e.g. wolf, cat, dog, even zebra all accumulate in a too small region, which is non-corresponding with the visual space. 
In contrast, all classes keep a reasonable distance in the rectified semantic space obviously.
This significantly proves the proposed semantic rectifying network is effective to help distinguish semantic features.
\section{Acknowledgements}
This work was supported by the National Natural Science Foundation of China (Nos. 61876121, 61472267,
61728205, 61502329, 61672371), Primary Research \& Development Plan of Jiangsu Province (No. BE2017663),
Aeronautical Science Foundation (20151996016), Jiangsu Key Disciplines of Thirteen Five-Year Plan(No. 20168765),
Suzhou Institute of Trade \& Commerce Research Project(KY-ZRA1805), 
and Foundation of Key Laboratory  in Science and Technology Development Project of Suzhou(No. SZS201609).
\section{Conclusion}
In this paper, we propose a novel generative approach
for Zero-Shot Learning (ZSL) by synthesizing visual features from rectified semantic features produced by a proposed semantic rectifying network (SRN). 
SRN maps original indiscriminative semantic features to rectified semantic features that are more distinguishable. 
Additionally, to guarantee the effectiveness of rectified semantic features and synthetic visual features, a pre-reconstruction and a post-reconstruction networks are proposed, and they preserve semantic details and keep the real visual distribution. 
Experimental results show that the proposed approach achieves state-of-the-art performance on ZSL task and boosts a great level(0.66\% $\sim$ 14.71\%) for GZSL.
{\small
	\bibliographystyle{IEEEbib}
	\bibliography{reference}
}
\end{document}